\pdfoutput=1

\documentclass[11pt]{article}
\usepackage{colortbl}

\usepackage{acl}
\usepackage{multirow}
\usepackage{xcolor}
\usepackage{arydshln}
\usepackage{bbding}
\usepackage{pifont}
\usepackage{amsmath}
\usepackage{subfig}
\usepackage{enumitem}
\usepackage[most]{tcolorbox}
\usepackage{array}

\usepackage{times}
\usepackage{latexsym}
\usepackage{booktabs}
\usepackage[T1]{fontenc}
\usepackage{soul}
\usepackage[utf8]{inputenc}
\usepackage{graphicx}
\usepackage{microtype}
\usepackage{inconsolata}
\usepackage{framed}

\definecolor{shadecolor}{gray}{0.95}
\definecolor{darkgreen}{HTML}{228B22}
\newcommand{\xmark}{\textcolor{red}{\ding{55}}}
\newcommand{\cmark}{\textcolor{darkgreen}{\ding{51}}}

\definecolor{softred}{rgb}{0.9, 0.2, 0.3} %
\definecolor{oceanblue}{rgb}{0.0, 0.5, 0.7} %
\definecolor{forestgreen}{rgb}{0.13, 0.55, 0.13} %

\DeclareRobustCommand{\llama}{%
  \begingroup\normalfont
  \raisebox{-0.4em}{%
  \includegraphics[height=11px]{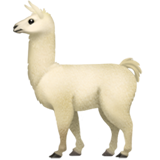}%
  }%
  \kern 0.4em%
  \endgroup
}
\DeclareRobustCommand{\gpt}{%
  \begingroup\normalfont
  \raisebox{-0.2em}{%
  \includegraphics[height=11px]{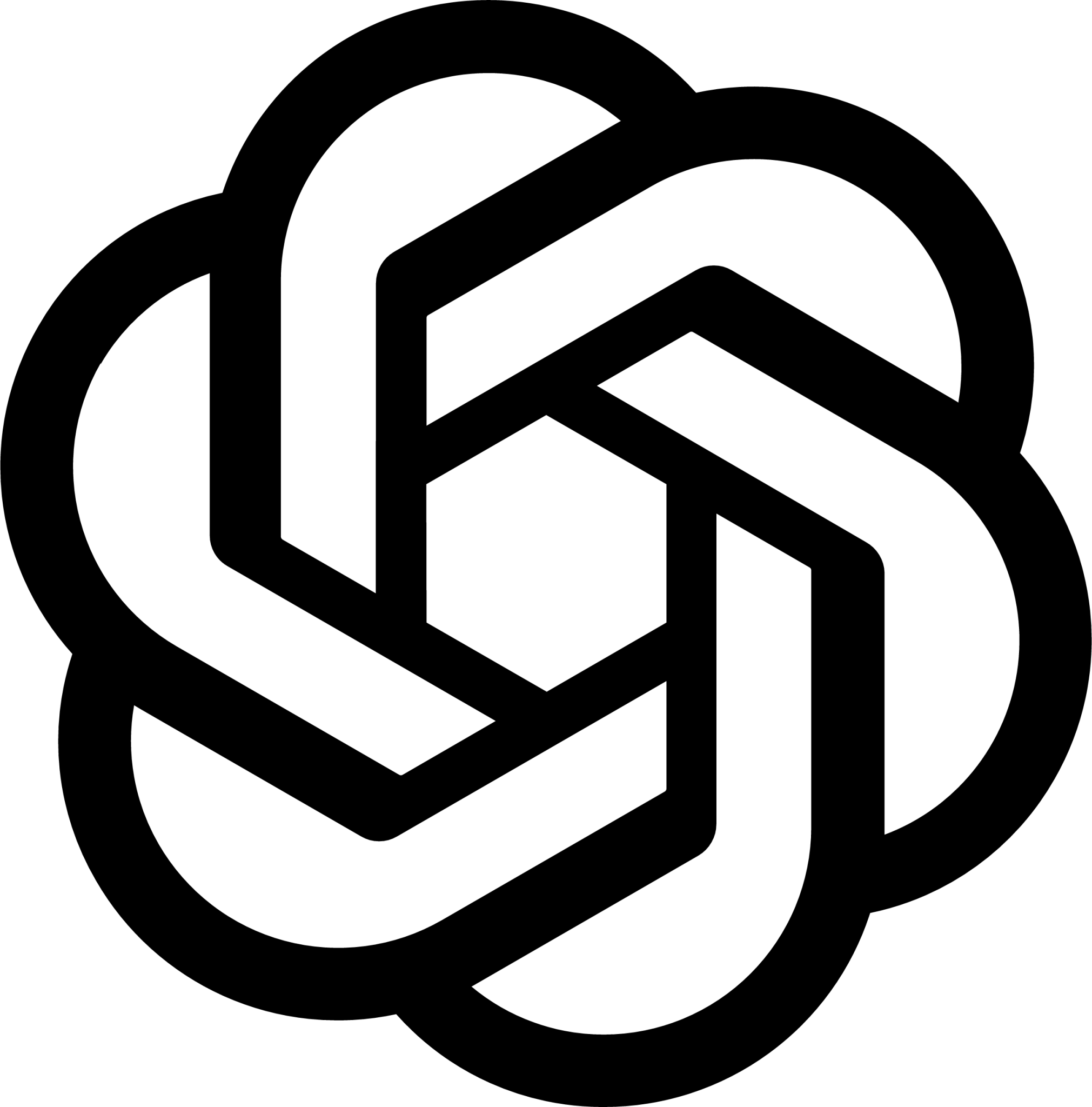}%
  }%
  \kern 0.4em%
  \endgroup
}
\DeclareRobustCommand{\mistral}{%
  \begingroup\normalfont
  \raisebox{-0.17em}{%
  \includegraphics[height=11px]{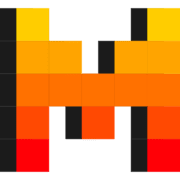}%
  }%
  \kern 0.4em%
  \endgroup
}

\title{AmbigNLG: Addressing Task Ambiguity in Instruction for NLG}

\author{Ayana Niwa$^{1,2,3}$\thanks{ ~~Work done at Megagon Labs \& Recruit Co., Ltd. Intellectual property rights retained by Megagon Labs \& Recruit Co., Ltd.}
  \quad
  Hayate Iso$^{1}$ \\
  $^1$ Megagon Labs \quad 
  $^2$ Recruit Co., Ltd. \quad 
  $^3$ MBZUAI \\
  {\tt ayana@megagon.ai} \quad
        {\tt hayate@megagon.ai} 
  }

\begin{document}
\maketitle
\begin{abstract}
We introduce AmbigNLG, a novel task designed to tackle the challenge of \textit{task ambiguity in instructions} for Natural Language Generation (NLG). 
Ambiguous instructions often impede the performance of Large Language Models (LLMs), especially in complex NLG tasks. 
To tackle this issue, we propose an ambiguity taxonomy that categorizes different types of instruction ambiguities and refines initial instructions with clearer specifications. 
Accompanying this task, we present AmbigSNI$_\texttt{NLG}$\footnote{AmbigSNI$_\texttt{NLG}$ dataset is available at \url{https://github.com/megagonlabs/ambignlg}}, a dataset consisting of 2,500 annotated instances to facilitate research on AmbigNLG. 
Through comprehensive experiments with state-of-the-art LLMs, we demonstrate that our method significantly enhances the alignment of generated text with user expectations, achieving up to a 15.02-point increase in ROUGE scores. 
Our findings highlight the importance of addressing task ambiguity to fully harness the capabilities of LLMs in NLG tasks.
Furthermore, we confirm the effectiveness of our method in practical settings involving interactive ambiguity mitigation with users, underscoring the benefits of leveraging LLMs for interactive clarification.
\end{abstract}

\section{Introduction}

Recent advancements in LLMs~\cite{brown2020language,touvron2023llama,jiang2023mistral} and instruction-tuning techniques~\cite{wei2022finetuned,sanh2022multitask,ouyang2022training,rafailov2023direct} have significantly expanded the capabilities of these models to tackle a wide range of problems through natural language interactions.
They now achieve near human-level performance on various benchmarks~\cite{hendrycks2021measuring,zheng2023judging}.
However, the effectiveness of LLMs is highly dependent on the clarity and specificity of the instructions they receive~\cite{wang2024promptagent}. 
Ambiguous instructions often lead to suboptimal or unintended results, highlighting a critical challenge in the practical deployment of these models.

Crafting precise instructions that unambiguously specify the expected outputs is inherently challenging for humans, especially for complex tasks such as Natural Language Generation (NLG). 
For instance, the instruction for summarization in the Super-Natural Instruction (SNI) benchmark~\cite{wang-etal-2022-super} is simply stated as ``\textit{Your task is to summarize them,}'' which is fairly ambiguous. 
It lacks crucial details such as the desired length of the summary, the key points to include, and the intended style. 
This type of ambiguity, known as \textit{task ambiguity}~\cite{NEURIPS2022_b43a0e8a}, is prevalent in various NLG tasks and must be addressed to effectively accomplish the task.

\begin{figure}[t]
    \centering
    \includegraphics[width=\linewidth]{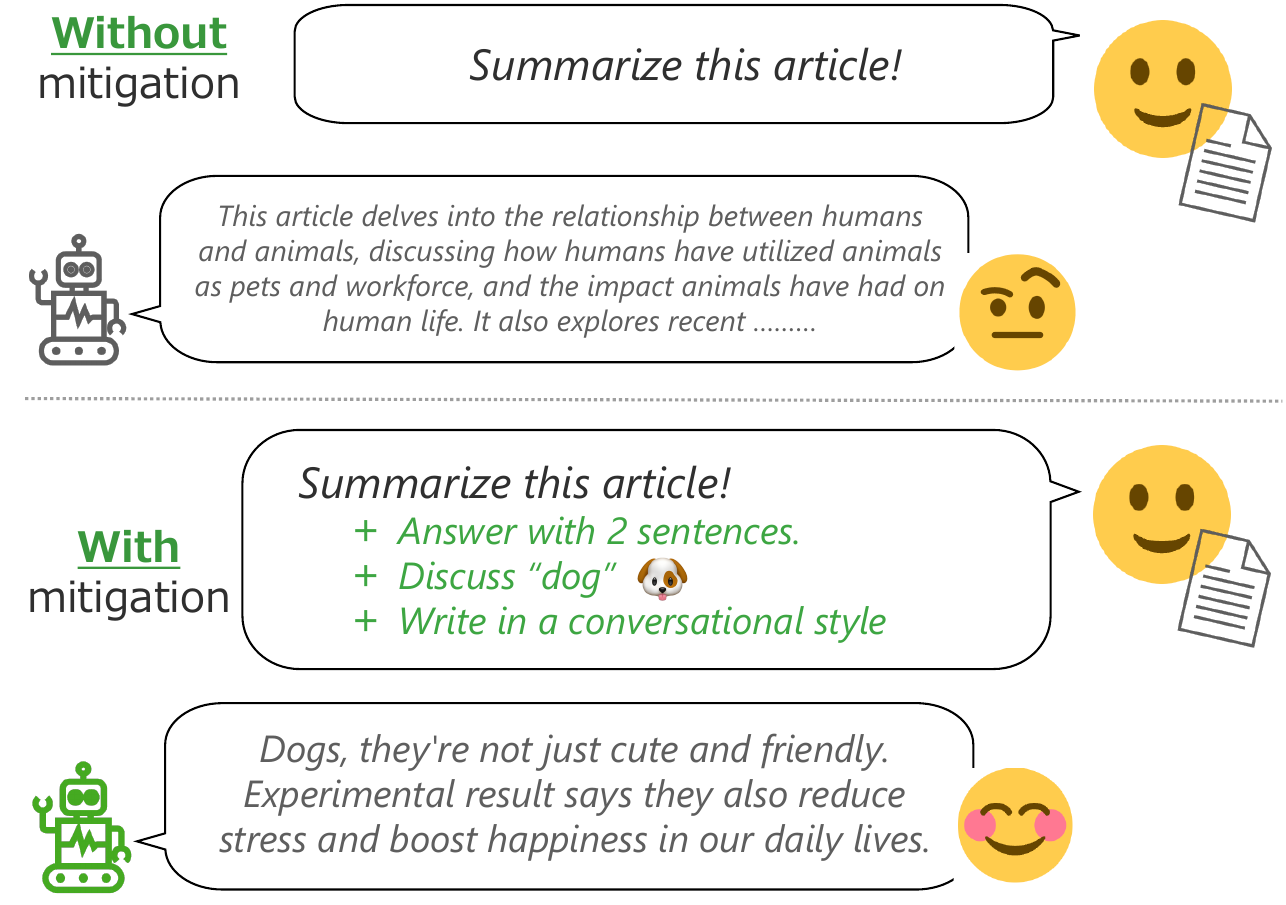}
    \caption{Overview of our mitigation approach for the AmbigNLG task. We address task ambiguity by incorporating {\color[HTML]{3EA642}{additional instructions}} into the initial instruction, thereby refining the task definition and improving the alignment of generated outputs with user expectations.}
    \label{fig:eyecatch}
\end{figure}

To address the issue of \textit{task ambiguity in instructions} for NLG, we first introduce \textbf{AmbigNLG}, a novel task aimed at identifying and mitigating ambiguities in various NLG instructions (\S\ref{sec:ambignlg}).
We then propose an ambiguity mitigation method that enhances initial instructions with clearer specifications (Figure~\ref{fig:eyecatch}, \S\ref{sec:mitigation}). 
This method involves establishing an ambiguity taxonomy to systematically categorize different types of instruction ambiguity in NLG tasks. 
Based on this taxonomy, we refine the initial instruction by appending additional instructions for each category.
This approach is intended for human-in-the-loop ambiguity mitigation, enabling users to directly choose the most suitable clarifications suggested by the LLM to effectively mitigate ambiguities (\S\ref{sec:e2e}).
Furthermore, to support our proposed method, we construct the AmbigSNI$_\texttt{NLG}$ dataset, comprising 2,500 instances annotated with ambiguity taxonomy and corresponding additional instructions (\S\ref{sec:dataset}).

We conducted a comprehensive analysis using several LLMs---including LLaMa-2, Mistral, Mixtral, and GPT-3.5---to evaluate the effectiveness of our proposed mitigation method.
The results indicate that our approach of providing additional instructions successfully mitigates task ambiguity, as evidenced by significant improvements in the alignment of generated text with user expectations, as well as a reduction in output diversity (\S\ref{sec:analysis}). 
Furthermore, a case study involving real human interaction confirms the practical utility, underscoring the importance of ambiguity mitigation in fully harnessing the capabilities of LLMs (\S\ref{sec:e2e}).

\begin{table*}[th]
    \centering
    \renewcommand{\arraystretch}{1.2}
    \small
    \resizebox{\linewidth}{!}{
    \begin{tabular}{p{1.5cm}p{4cm}p{6cm}p{4cm}}
        \toprule
        \textbf{Taxonomy} & \textbf{Definition} & \textbf{Template} &\textbf{Example of Filler} \\ \midrule
     \rowcolor{gray!15}    \textsc{Context}& Uncertainty of the situation or background & \textbf{\texttt{Additional context:  \_\_\_}}& The main factors of climate change are natural phenomena and human activities.\\
     \textsc{Keywords}& Not sure which words to include & \textbf{\texttt{Include \_\_\_ in your response.}} &global warming\\
     \rowcolor{gray!15}\textsc{Length}& Underspecified length & \textbf{\texttt{Answer with \_\_\_ words.} }& 10 to 20 \\
     \textsc{Planning}&Uncertainty of the text structure & \textbf{\texttt{Please generate the output based on the following outline: 1. \_\_\_ 2.\_\_\_, ...}} &1. a brief definition, 2. causes, ...\\
     \rowcolor{gray!15}  \textsc{Style}& Underspecified writing style&  \textbf{\texttt{Write in a \_\_\_ style.} }& persuasive \\
     \textsc{Theme} & Uncertainty of the main subject & \textbf{\texttt{Primarily discuss the following theme: \_\_\_.}} &the impact of human activities\\
       \bottomrule
    \end{tabular}
           }
    \caption{Ambiguity taxonomy, definitions, templates, and examples of fillers for each template. The filler serves as an example given the instruction `Write a summary about climate change. This taxonomy helps in systematically categorizing and addressing different types of ambiguities in NLG tasks.\label{tab:ambig_category}}
\end{table*}

\section{Task: AmbigNLG}
\label{sec:ambignlg}
We address the challenge of task ambiguity in instruction, which arises from insufficiently defined tasks.
Our aim is to enhance the accuracy of text generation to better meet users' expectations.
Unlike previous studies that focus on ambiguity in Natural Language Understanding (NLU) tasks~\cite{finn2018probabilistic,NEURIPS2022_b43a0e8a,tamkin2023task}, our work uniquely concentrates on mitigating ambiguity in NLG task instructions.
In the NLG setting, addressing ambiguity requires more adaptable strategies due to the multifaceted nature of ambiguities, such as summary length and content.
To this end, we propose AmbigNLG task, specifically designed to tackle task ambiguity in NLG instructions.

\subsection{Problem Definition}
\label{sub:definition}
In instruction-based NLG tasks, the goal is to generate an output text $y$ from a given input text $x$, following an instruction $I$~\cite{wei2022finetuned,wang-etal-2022-super}.
For a specific input $x$ and instruction $I$, there often exists a range of valid output texts, denoted as $\mathcal{Y}_{\text{valid}}$.
Modern NLG models such as LLMs are capable of generating such valid outputs $\hat{y} \in \mathcal{Y}_{\text{valid}}$.
However, if the instruction $I$ is not well specified, the LLMs may generate an output that, while valid $\hat{y} \in \mathcal{Y}_{\text{valid}}$, does not align with the user's actual intent---that is, $\hat{y} \not\in\mathcal{Y}_{\text{desired}}$, where $\mathcal{Y}_{\text{desired}} \subseteq \mathcal{Y}_{\text{valid}}$.
We define this phenomenon as \textbf{\textit{task ambiguity in instructions}} for NLG, referring to unclear or insufficiently detailed instructions that hinder the LLM's ability to generate text aligned with user intentions.
Conversely, if the set of valid outputs $\mathcal{Y}_{\text{valid}}$ matches the user's desired outputs $\mathcal{Y}_{\text{desired}}$, the instruction $I$ is considered unambiguous for the input $x$.

\subsection{Task Ambiguity Mitigation}
\label{sec:definition}
Building on the definition above, we formulate \textit{task ambiguity \ul{mitigation} in instructions} as the process of refining an initial instruction $I_{\text{init}}$ into a more precise instruction $I_{\text{refined}}$.
This refinement aims to narrow the set of valid output texts $\mathcal{Y}_{\text{valid}}$ to more closely align with the user's desired outputs $\mathcal{Y}_{\text{desired}}$.
Given the intractable nature of defining both valid and desired output sets, we simplify the problem by using a reference text $y_{\text{ref}}$ as a proxy for the desired output.
The objective is to refine the initial instruction so that the generated text $\hat{y}$ more closely matches the reference text $y_{\text{ref}}$.

\section{Method for Ambiguity Mitigation\label{sec:mitigation}}

\subsection{Ambiguity Taxonomy}
\label{sub:taxonomy}
To effectively mitigate task ambiguity in instructions, it is crucial to first identify and understand the types of ambiguities present in instruction-based NLG datasets. 
To this end, we conducted a comprehensive literature survey to explore the fundamental components in NLG systems~\cite{reiter1997building,mcdonald-pustejovsky-1985-computational,kukich-1983-design,barzilay-lapata-2005-modeling,reitter-etal-2006-computational,fan-etal-2018-controllable}.
Building upon insights from the literature, we manually analyzed 100 instruction-based NLG instances from Super-Natural Instruction (SNI) benchmark~\cite{wang-etal-2022-super} to build an ambiguity taxonomy.\footnote{
Specifically, each instance consists of a triplet: input, output, and instruction, with a total of 23,796 words across 100 randomly sampled instances.
After comparing these triplets with a broad range of NLG literature and thorough detailed discussions, we identified 484 specific ambiguous points and categorized them.}
This analysis led us to identify six dominant types of task ambiguity: \textsc{Context}, \textsc{Keywords}, \textsc{Length}, \textsc{Planning}, \textsc{Style}, and \textsc{Theme}, as detailed in Table~\ref{tab:ambig_category}.

\subsection{Instruction Refinement}
\label{sub:instruction_refining}
To mitigate task ambiguity in instructions, we refine the initial instruction using our proposed taxonomy.
Directly rewriting the initial instruction $I_{\text{init}}$ to craft a refined instruction $I_{\text{refined}}$ presents challenges in maintaining consistency and quality.
Therefore, we simplify the process by appending \textit{additional instructions} $\{I_{c_1}, \dots, I_{c_n}\}$  to address each identified task ambiguity category $\{c_1, \dots, c_n\}$ found in the initial instruction $I_{\text{init}}$.
We concatenate these additional instructions with the initial instruction to create the refined instruction $I_{\text{refined}} = I_{\text{init}} \oplus I_{c_1} \oplus \dots \oplus I_{c_n}$, where $\oplus$ denotes the text concatenation operator. 
These refined instructions $I_{\text{refined}}$ serve as pseudo-references for unambiguous instructions, facilitating the study of ambiguity mitigation in NLG tasks.\footnote{If multiple ambiguities are present, the additional instructions are concatenated in alphabetical order based on the ambiguity category names.}

\section{Dataset: AmbigSNI$_\texttt{NLG}$\label{sec:dataset}}

To evaluate our mitigation method described in \S\ref{sec:mitigation}, we constructed the AmbigSNI$_\texttt{NLG}$ dataset.\footnote{Details on intended usage are provided in Appendix~\ref{app:dataset_usage}.}
AmbigSNI$_\texttt{NLG}$ is derived from the NLG dataset within the SNI benchmark, which encompasses 1,616 diverse NLP tasks.
Each instance in this dataset is annotated with our ambiguity category $c$ and the corresponding additional instruction $I_{c}$.

\begin{figure*}[t]
    \centering
    \includegraphics[width=1\linewidth]{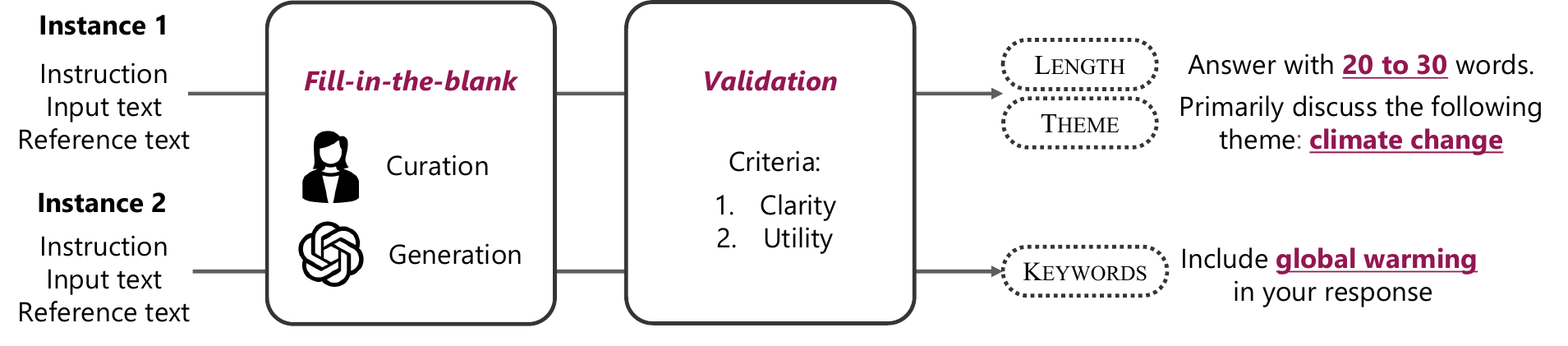}
    \caption{Dataset creation process. The process includes curating high-quality manual annotations, generating additional instruction candidates, and validating these candidates to ensure clarity and utility. }
    \label{fig:data_creation_proess}
\end{figure*}

\begin{table*}[ht]
    \centering
    \small
        \setlength{\tabcolsep}{3pt}
    \begin{tabular}{lrrrrrrrrrrrr}
    \toprule
    && \multicolumn{5}{c}{\# Additional Instructions \%} & \multicolumn{6}{c}{\# Ambiguity Type \%} \\
    \cmidrule(l){3-7} \cmidrule(l){8-13}
   \textbf{Split} & \#data & 0 & 1 & 2 & 3 & 4+ & \textsc{context} & \textsc{keywords} & \textsc{length} & \textsc{planning} & \textsc{style} & \textsc{theme} \\
    \midrule
    Demonstration & 500 & 25.6 & 33.4 & 26.2 & 11.4 & 3.4 & 35.2 & 39.4 & 19.8 & 7.0 & 2.0 & 30.4 \\
    Evaluation & 2,000 & 27.8 & 35.8 & 21.6 & 11.6 & 3.2 & 34.6 & 38.6 & 18.5 & 5.9 & 1.6 & 27.7 \\
    \bottomrule
    \end{tabular}
    \caption{Data statistics. Percentage of ambiguity categories assigned to each instance (\# Additional Instructions), and percentage of instances assigned to each category (\# Ambiguity Type).}
    \label{tab:num_category}
\end{table*}

\subsection{LLM-in-the-loop Annotation}
\label{sub:annotation}

Annotating ambiguity categories $c$ and additional instruction $I_c$ through crowdsourcing is challenging due to the open-ended nature of text generation tasks.
To address this issue, we adopt an LLM-in-the-loop approach, where we manually \textit{curate} and \textit{verify} the dataset by guiding the LLM's generation~\cite{ding-etal-2023-gpt,gilardi2023chatgpt,zhang-etal-2024-xatu}.
 To ensure consistency in the annotation of additional instructions, we developed specific templates $t_c$ for each ambiguity category $c$, as shown in Table~\ref{tab:ambig_category}.
These templates are filled out to create the additional instructions~\cite{iso-etal-2020-fact,liu-etal-2023-improving,zhou2023instructionfollowing,iso-2024-autotemplate-simple}.
The data creation process is outlined in Figure~\ref{fig:data_creation_proess}.
Note that for the \textsc{Keywords} and \textsc{Length}, additional instructions can be curated using a rule-based approach described in Appendix~\ref{app:rule_based_annotation}; therefore, only validation is performed for these categories.

\paragraph{Curation}
We curated high-quality manual annotations of existing ambiguity in instructions.
First, we manually analyzed 100 instruction-based NLG instances from SNI benchmark to identify types of ambiguities. 
These 100 samples were randomly selected to cover a wide variety of tasks, including question answering, summarization, and dialogue generation.
Then, we annotated the additional instructions for each instance by filling in the blanks of the corresponding templates.
To ensure the quality of the additional instructions, we employed a rigorous annotation process detailed in \S\ref{app:curation}.

\paragraph{Generation}
Given the manual annotations, we fine-tuned GPT-3.5 to generate the additional instruction for each ambiguity category.\footnote{\url{https://openai.com/blog/gpt-3-5-turbo-fine-tuning-and-api-updates}}
We provided initial instruction $I_{\text{init}}$, input text $x$, reference text $y_{\text{ref}}$, and the template $t_{c}$ corresponding to each ambiguity category $c$ as inputs to generate the additional instruction candidates $\hat{I}_{c}$ for all categories.\footnote{We minimized information leakage from the reference text by carefully designing the prompt. Details, analysis, and the prompt are described in the Appendix.}

\paragraph{Validation}
\label{validation}
Finally, we validate the generated additional instruction candidates $\hat{I}_{c}$ and retain only those that meet the following criteria to obtain  the final additional instructions $I_c$:
\begin{itemize}[noitemsep, nolistsep]
    \item \textbf{Clarity}: We assess whether the candidate $\hat{I}_c$ enhances the clarity of the initial instruction $I_{\text{init}}$. To facilitate scalability, we employ GPT-4 as an evaluator.\footnote{Our in-house evaluation showed that GPT-4's assessment aligned with human judgments in 91\% of cases.} Only additional instructions that reduce ambiguity in the initial instruction are accepted under this criterion. 
    \item \textbf{Utility}: We determine whether the generated candidate $\hat{I}_c$ helps generate output text that more closely aligns with the desired output. Specifically, we compare the ROUGE-L F1 scores  of outputs generated before and after appending $\hat{I}_c$ to $I_{\text{init}}$, resulting in the refined instruction  $\hat{I}_{\text{refined}}:= I_{\text{init}} \oplus \hat{I}_{c}$. Using GPT-4, we generate 20 output samples for both $I_{\text{init}}$ and $\hat{I}_{\text{refined}}$. We then perform statistical significance testing to evaluate whether the inclusion of $\hat{I}_c$ leads to output $\hat{y}$ that is significantly closer to the reference text $y_{\text{ref}}$. Only additional instructions demonstrating a significant improvement are retained. 
\end{itemize}

\subsection{Dataset Statistics\label{sec:dataset_stats}}
\begin{figure*}
    \centering
    \includegraphics[width=\linewidth]{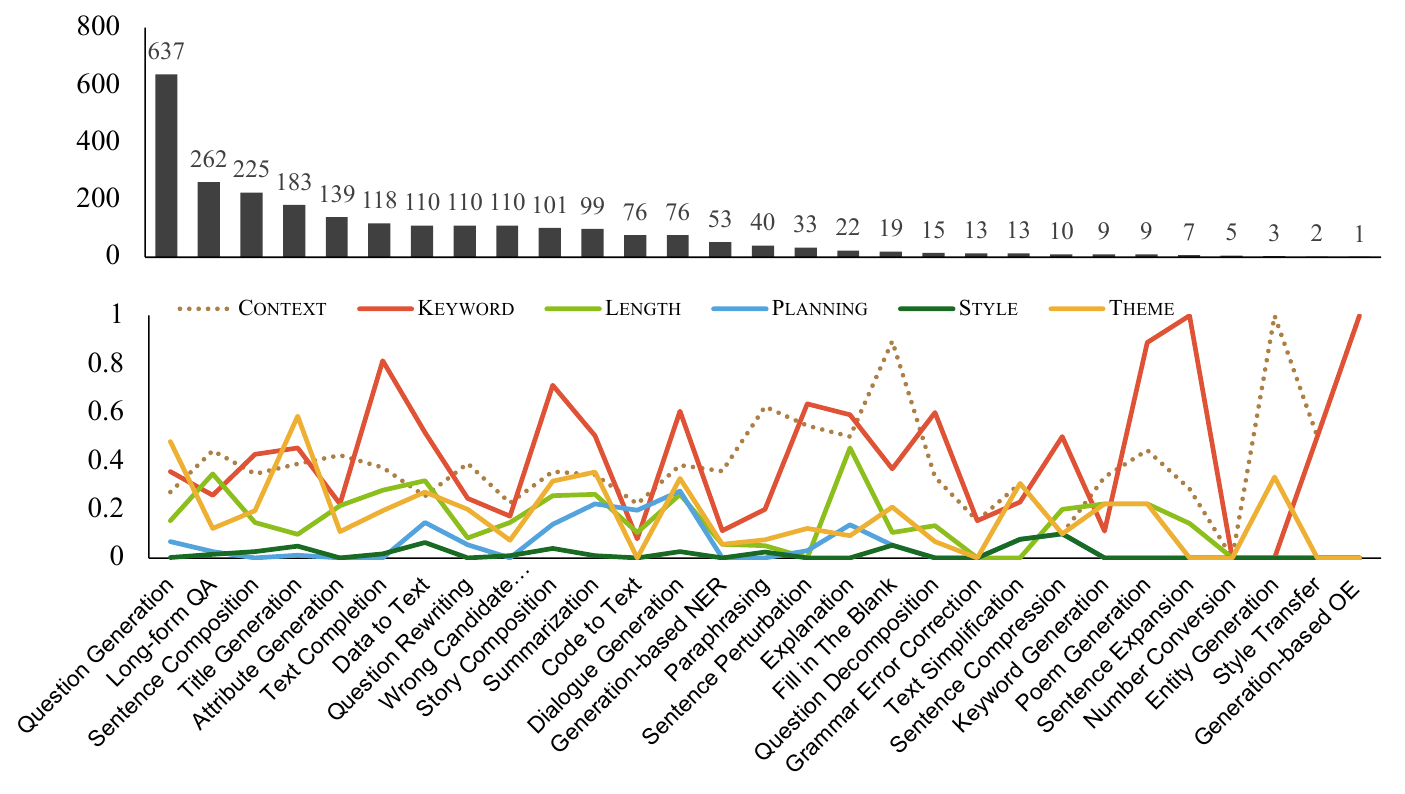}
    \caption{Distributions of the dataset. The upper bar graph displays the number of instances per task, while the lower line graph shows the proportion of instances assigned to ambiguous categories for each task.}
    \label{fig:stats}
\end{figure*}

The AmbigSNI$_\texttt{NLG}$ dataset comprises 2,500 meticulously curated instances covering a wide range of NLG tasks as illustrated in Figure~\ref{fig:stats}.
The dataset is randomly split into 2,000 instances for evaluation and 500 for demonstrations.
As shown in Table~\ref{tab:num_category}, approximately 75\% of the instances present at least one category of task ambiguity in instructions, and around 35\% contain multiple types of ambiguities.

Our dataset reveals a significant prevalence of categories such as \textsc{Context}, which encompasses background information about the task and necessary knowledge; \textsc{Keywords}, which specifies words that should be included; and \textsc{Theme} which pertains to information about the content.
This indicates that these aspects are particularly susceptible to ambiguity in NLG task instructions.

When analyzing the statistics for each task, Question generation is the most populated task, followed by Long-form QA, Sentence Compression, and Title Generation.
Tasks requiring consideration of multiple topics---such as Question Generation, Title Generation, and Summarization---are predominantly associated with the \textsc{Theme} category.
In contrast, tasks like Code to Text, designed to preserve content fidelity, exhibit a generally lower frequency of ambiguity categories except for \textsc{Context}.
See examples and additional statistics in the Appendix.

\section{Experiments}
\label{sec:analysis}

In this section, we empirically assess the effectiveness of our annotated additional instructions presented in \S\ref{sec:dataset} in mitigating the \textit{task ambiguity in instructions} defined in \S\ref{sec:definition}. 
Specifically, the goal of this section is to verify whether the model can utilize these additional instructions to mitigate ambiguities effectively.

\subsection{Settings}
\paragraph{Methods}
We evaluate two approaches for constructing refined instructions $I_{\text{refined}}$. 
The first approach, referred to as \texttt{Taxonomy}, involves concatenating our annotated additional instructions $\{I_{c_1}, \dots, I_{c_n}\}$ to the initial instruction $I_{\text{init}}$. 
Formally, the refined instruction is given by: $I_{\text{refined}}:= I_{\text{init}} \oplus I_{c_1} \oplus \dots \oplus I_{c_n}$.\footnote{We evaluated whether increasing instruction complexity by concatenating additional instructions affects the instruction-following capability of LLMs. Our experiment showed that this treatment does not impact their ability. See more details in the Appendix.}

The second approach, termed \texttt{Generic}, constructs the refined instruction by appending a generic additional instruction $I_{\text{generic}}$ to the initial instruction $I_{\text{init}}$: $I_{\text{refined}}:=I_{\text{init}}\oplus I_{\text{generic}}$.
This method serves as a baseline to evaluate the importance of our ambiguity taxonomy in mitigating ambiguity. Specifically, we employed the same generation pipeline described in \S\ref{sec:dataset}, but used a generic template, `Additional information: \_\_\_\_,' to create the additional instruction $I_{\text{generic}}$.\footnote{For instance, in a summarization task, the generic additional instruction might be, ``Please make sure to include the main points of the passage in your summary, even if they need to be slightly adjusted for conciseness.''}

\paragraph{Models}
We employ instruction fine-tuned LLaMA-2 (\llama; 7B)~\cite{touvron2023llama}, Mistral (\mistral; 7B)~\cite{jiang2023mistral} and Mixtral (\mistral; 8x7B)~\cite{jiang2024mixtral} for open-sourced LLMs.
Additionally, we utilize GPT-3.5 (\gpt; n/a) as a proprietary model.\footnote{We exclude GPT-4 from our experiments as it serves as a data generator.} 
To optimize space in our tables, each model is represented by an emoji along with its parameter size as an identifier.
 
\paragraph{Metrics}

To quantify the effect of task ambiguity mitigation in instructions on LLMs' responses, we measure two key aspects: \texttt{Alignment} and \texttt{Focus}.

For \texttt{Alignment}, we assess how well the LLMs generate responses that align with the user's expectations, as represented by the reference text $y_{\text{ref}}$, when additional instructions are provided. 
This is measured by the relative gains in reference-based metrics, specifically ROUGE-L and BERTScore.\footnote{\texttt{distilbert-base-uncased} with baseline re-scaling.}
We compare the outputs generated using only initial instructions $I_{\text{init}}$ with those generated using the refined instructions $I_{\text{refined}}$.

For \texttt{Focus},we evaluate the extent to which ambiguity mitigation narrows the output space $\mathcal{Y}_{\text{valid}}$. 
Our hypothesis is that effective ambiguity mitigation will result in less diverse outputs. 
To quantify this, we compute the ROUGE-L score for each pair of sampled responses and average these scores, defined as the Intra-RL score~\cite{pmlr-v97-shen19c,iso-etal-2022-comparative}: $\frac{2}{N (N-1)} \sum_{j < k}\text{ROUGE-L}(\hat{y}_{j}, \hat{y}_{k})$, where $N$ is the number of sampled responses.
A higher Intra-RL score indicates that the sampled responses are more similar to each other, suggesting a narrower output space. 
We report the relative gains of Intra-RL scores to quantify the improvement, comparing outputs from initial instructions to those from refined instructions.

For these evaluations, we sample 20 responses per instance using a \texttt{temperature} setting of 1.0.

\begin{table}[t]
    \centering
    \resizebox{\linewidth}{!}{
    \small
    \begin{tabular}{cccccc}
    \toprule
    & & & \multicolumn{2}{c}{\texttt{Alignment}} & \texttt{Focus}\\\cmidrule(l){4-5}\cmidrule(l){6-6}
    \textbf{Model} & \# Param & Method & RL & BS & IntraRL \\\midrule
    \multirow{2}{*}{\includegraphics[width=18px]{img/llama.png}}& \multirow{2}{*}{7B} & \texttt{Generic} & 0.44 & 1.13 & -0.09 \\
    &  & \texttt{Taxonomy} & 7.96 & 9.08 & 0.16\\\midrule
    \multirow{4}{*}{\includegraphics[width=18px]{img/mistral.png}} & \multirow{2}{*}{7B} & \texttt{Generic} & 0.14 & 0.59 & -0.08\\
    &  & \texttt{Taxonomy} & 6.83 & 7.78 & 0.25\\
    \cmidrule(l){2-6}
    & \multirow{2}{*}{8x7B} & \texttt{Generic} & 0.46 & 1.14 & -0.09\\
    &  & \texttt{Taxonomy} & 8.56 & 9.16 & 0.30\\\midrule
    \multirow{2}{*}{\includegraphics[width=18px]{img/gpt.png}} & \multirow{2}{*}{n/a} & \texttt{Generic} & 1.47 & 1.69 & 0.16 \\
    &  & \texttt{Taxonomy} & 15.02 & 13.62 & 0.66\\
    \bottomrule
    \end{tabular}
    }
    \caption{Relative gains in performance metrics for ambiguity mitigation. The table shows the relative gains in ROUGE-L (RL), BERTScore (BS), and Intra-RL for different models and methods. }
    \label{tab:gen}
\end{table}

\subsection{Results\label{sub:ex1}}
Table~\ref{tab:gen} presents the relative gains in ROUGE-L, BERTScore, and Intra-RL metrics for ambiguity mitigation across different models.
For \texttt{Alignment}, the results demonstrate substantial improvements in both ROUGE-L and BERTScore when using the \texttt{Taxonomy} compared to the \texttt{Generic}. 
Specifically, GPT-3.5 exhibits the highest gains, with a 15.02-point increase in ROUGE-L and a 13.62-point increase in BERTScore. 
This significant enhancement indicates that the \texttt{Taxonomy} effectively aligns generated responses with user expectations.

For \texttt{Focus}, the Intra-RL scores reveal that the \texttt{Taxonomy} consistently narrows the output space more effectively than the \texttt{Generic}. 
For instance, GPT-3.5 shows a significant gain of 0.66 in Intra-RL, while LLaMA-2 (7B) and Mistral (7B, 8x7B) also demonstrate positive gains of 0.16, 0.25, and 0.30, respectively.
This suggests that the \texttt{Taxonomy} approach reduces variability in the generated outputs, focusing more closely on the desired content.

Overall, the results highlight that the \texttt{Taxonomy} outperforms the baseline without ambiguity mitigation and the \texttt{Generic} method.
It not only improves alignment with user expectations but also effectively narrows the output space, thereby mitigating task ambiguity in instructions for LLMs.\footnote{Further analysis is provided in Appendix~\ref{app:further_analysis}.}

\begin{figure}[t]
    \centering
    \includegraphics[width=0.9\linewidth]{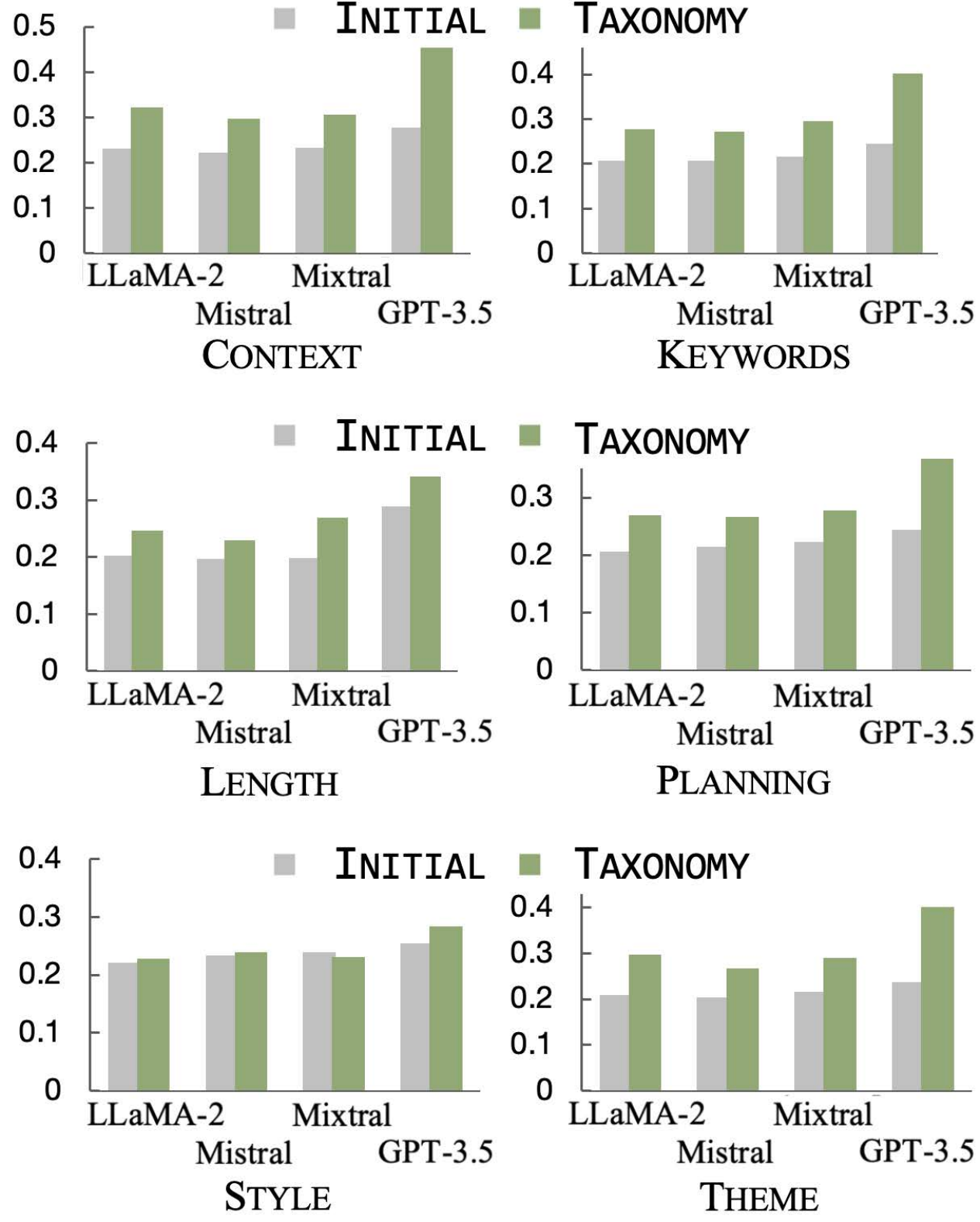}
    \caption{Mitigation results for each taxonomy.}
    \label{fig:exp1_taxonomy}
\end{figure}

\begin{figure}[t]
    \centering
    \includegraphics[width=\linewidth]{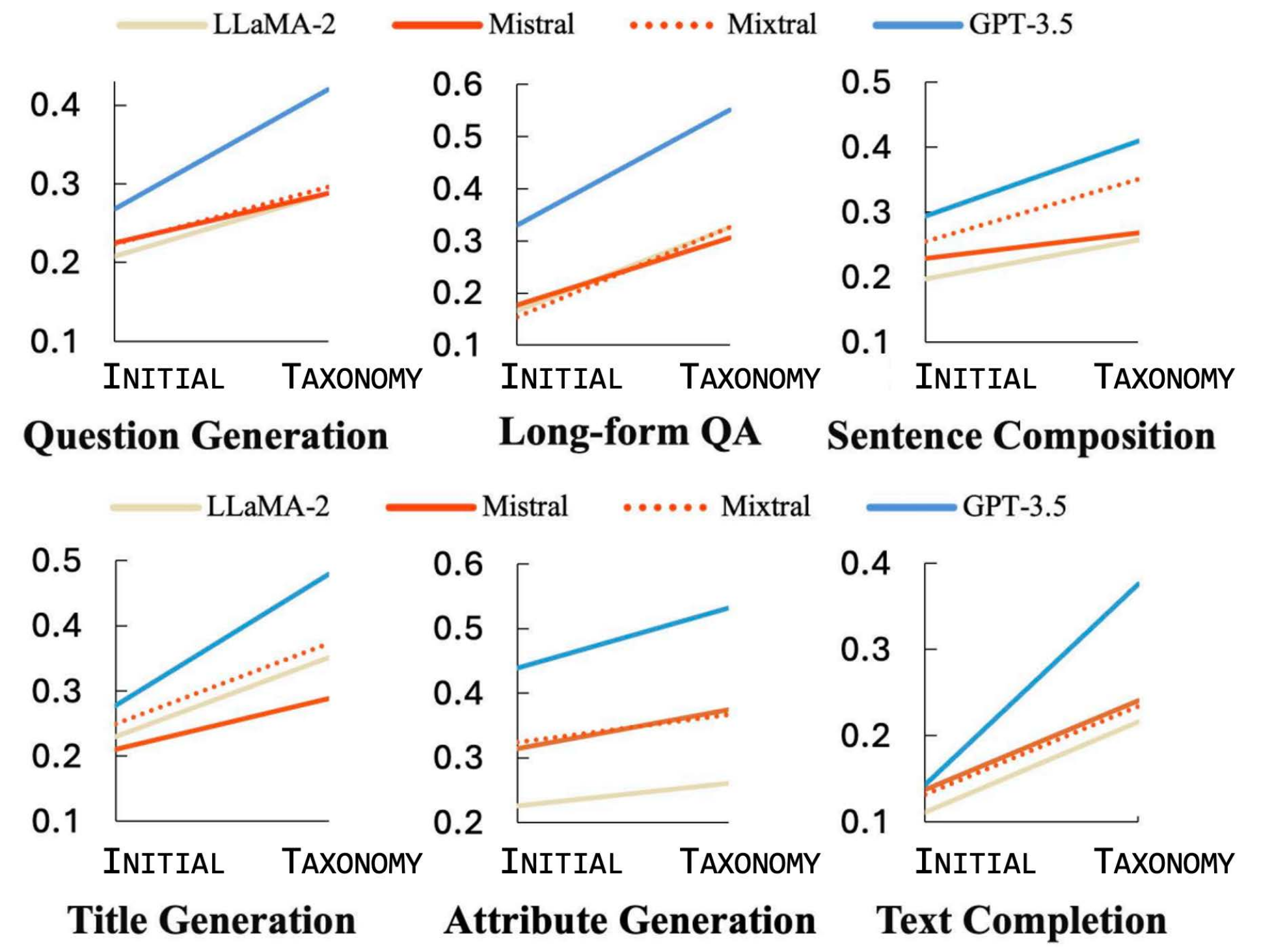}
    \caption{Mitigation results across the top-6 most frequent tasks in AmbigSNI$_{\texttt{NLG}}$. The figure demonstrates that ambiguity mitigation consistently enhances performance across different NLG tasks, as indicated by the ROUGE-L score improvements.}
    \label{fig:exp1_tasks}
\end{figure}

\paragraph{Analysis\label{para:analysis}}
We provide additional insights into the effectiveness of the \texttt{Taxonomy} method across different ambiguity categories and NLG tasks.
Figure~\ref{fig:exp1_taxonomy} illustrates the improvements in ROUGE-L scores across all categories and nearly all models.
Notably, categories directly related to the content, such as \textsc{Context}, \textsc{Keywords}, and \textsc{Theme}, show substantial improvements. 
This underscores the importance of an ambiguity taxonomy and explicit, category-specific additional instructions for effective ambiguity mitigation.
Figure~\ref{fig:exp1_tasks} presents the ROUGE-L score improvements for various NLG tasks, demonstrating that ambiguity mitigation consistently enhances performance regardless of the task type. 
This highlights the significance of ambiguity mitigation in fully leveraging the capabilities of LLMs for diverse NLG tasks.
The comprehensive results are presented in Table~\ref{tab:all_tasks}.

\section{Human-in-the-loop Ambiguity Mitigation}\label{sec:e2e}
\begin{figure}[ht]
    \centering
    \includegraphics[width=\linewidth]{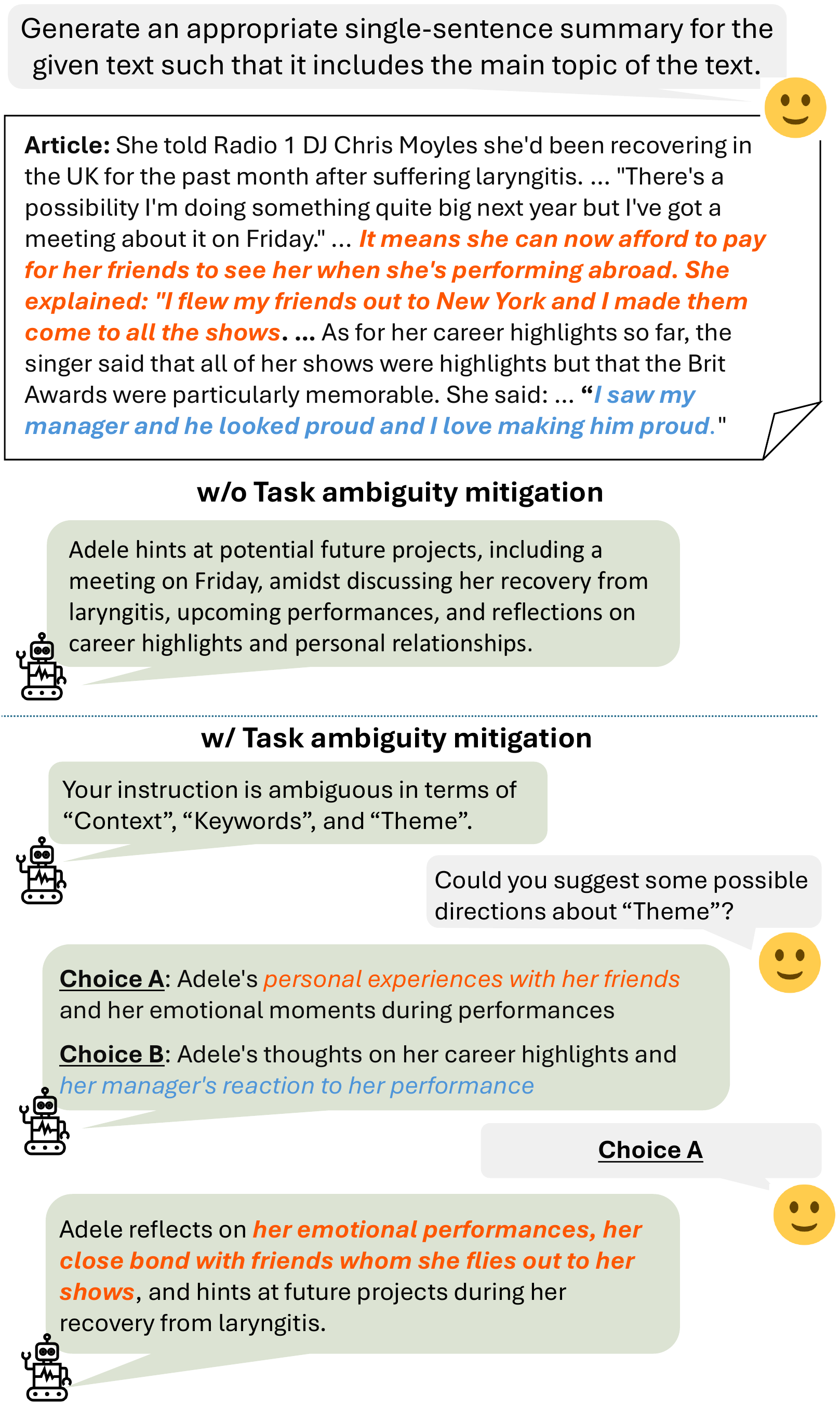}
    \caption{Example with pipeline mitigation.}
    \label{fig:e2e}
\end{figure}

To assess the practical utility of our proposed ambiguity mitigation framework, we conducted a case study involving human interaction. 
This experiment aims to assess whether LLM-generated additional instructions can effectively guide the generation of desired outputs in real-world scenarios.

\subsection{Experimental Design}

As illustrated in Figure~\ref{fig:e2e}, our case study is designed to simulate real-world scenarios in which users engage with LLMs to clarify ambiguous instructions. The goal is to improve the alignment of the generated outputs with user intent through the following steps:
\begin{enumerate}
    \item Given an initial instruction $I_{\text{init}}$, the LLM identifies a potential ambiguity $c$ (\S\ref{sub:ex2}) and suggests additional instructions $\{\hat{I}_{c}^1, \dots, \hat{I}_{c}^N\}$ to address these ambiguities (\S\ref{sub:ex3}).
    \item The user then selects the most appropriate additional instruction provided by the LLM to mitigate the ambiguities in $I_{\text{init}}$.
    \item Finally, the LLM generates the output based on the refined instruction $I_{\text{refined}}$ (\S\ref{sub:ex14}).
\end{enumerate}

\subsection{Identifying Ambiguity in Instructions}
\label{sub:ex2}

We begin by investigating the ability of LLMs to identify task ambiguity in instructions, framing this as a binary classification problem for each ambiguity category.

\paragraph{Settings}

Experiments were conducted in both zero-shot and in-context settings. In the in-context setting, we retrieved 8 similar examples from the demonstration set using \texttt{all-mpnet-base-v2} as the retriever and incorporated these examples along with their labels into the context provided to the LLMs. 
To address the imbalance in the distribution of ambiguity labels, we evaluated the models using True Positive Rate (TPR), True Negative Rate (TNR), and accuracy (Acc). Additionally, we used exact match accuracy (EM) to assess the overall success in identifying all ambiguity labels.

\paragraph{Results}

Table~\ref{tab:cls} illustrates that in zero-shot settings, all LLMs tended to classify instructions as ambiguous, resulting in high TPR but low TNR and consequently near-zero EM scores. 
However, with in-context demonstrations, all open-sourced LLMs exhibit a more balanced evaluation of ambiguity, leading to higher Acc and EM. 
This indicates that in-context demonstrations, rather than model size, play a crucial role in accurately identifying task ambiguity. 
Interestingly, GPT-3.5 did not  follow this trend, implying it may prioritize its own decision over the influence of in-context demonstrations.

\begin{table}[t]
    \centering
    \small
    \resizebox{\linewidth}{!}{
    \begin{tabular}{cccrrrr}
    \toprule
    && & \multicolumn{3}{c}{Category Average}   & All    \\
    \cmidrule(lr){4-6} \cmidrule(lr){7-7}
    \textbf{Model}  & \#Param        & ICL &   TPR &   TNR &   Acc &    EM \\
    \midrule
    \multirow{2}{*}{\includegraphics[width=18px]{img/llama.png}} & \multirow{2}{*}{7B} & \xmark & 98.31 &  1.82 & 22.07 &  0.10 \\
     &  & \cmark & 14.01 & 86.81  & 70.83 & 20.20 \\\midrule
    \multirow{4}{*}{\includegraphics[width=18px]{img/mistral.png}}& \multirow{2}{*}{7B} & \xmark & 99.93 &  0.15  & 21.27 &  0.00 \\
     &  & \cmark & 55.01 & 49.31  & 50.15 & 13.30 \\
     \cmidrule(l){2-7}
     & \multirow{2}{*}{8x7B} & \xmark & 96.59 &  4.38  & 23.70 &  0.10 \\
     &  & \cmark & 20.14 & 82.85  & 70.04 & 23.15 \\\midrule
    \multirow{2}{*}{\includegraphics[width=18px]{img/gpt.png}} & \multirow{2}{*}{n/a} & \xmark & 79.66 & 23.00 &  34.63 &  3.55 \\
     & & \cmark & 87.48 & 10.80  & 26.92 &  1.30 \\
    \bottomrule
    \end{tabular}
    }
    \caption{Performance of ambiguity identification. The table shows the True Positive Rate (TPR), True Negative Rate (TNR), accuracy (Acc), and exact match accuracy (EM) for identifying task ambiguity across different models and settings.}
    \label{tab:cls}
\end{table}

\subsection{Suggesting Addition Instructions}
\label{sub:ex3}

We next evaluate the ability of LLMs to generate useful additional instructions for mitigating task ambiguity. 
Specifically, we investigate whether LLMs can suggest suitable options for an additional instruction $I_c$ based on the identified ambiguity category $c$, allowing users to choose the most appropriate one.

\paragraph{Settings}

We employed templates specific to each ambiguity category to generate candidates by either sampling or batching $N$ suggestions simultaneously. 
We framed this suggestion task as a recommendation problem, assessing the candidates based on their \textbf{Relevance} and \textbf{Diversity}. 
For \textbf{Relevance}, we measured the highest ROUGE-L score (RL@$N$) and semantic similarity (Para@$N$) between the generated candidates and the reference $I_c$ in AmbigSNI$_\texttt{NLG}$. 
For \textbf{Diversity}, we calculated the Intra-RL score among the candidates to assess the variety of the suggestions.

\paragraph{Results}

Table~\ref{tab:suggest} presents the efficacy of LLMs in suggesting additional instructions to mitigate ambiguity when $N = 10$.
The results indicate that for LLaMA-2, Mistral, and Mixtral, generating more diverse outputs leads to higher surface-level and semantic similarity with the reference $I_c$, confirming the benefit of generating varied suggestions to address ambiguity. 
Conversely, for GPT-3.5, enhancing diversity through batch generation significantly decreases relevance, indicating that while GPT-3.5 excels at generating optimal additional instructions, forcing it to generate diverse outputs can impair this capability. 
 This underscores the importance of tailoring generation settings to each model's strengths.
 
\begin{table}[t]
    \centering
    \small
    \setlength{\tabcolsep}{3pt}
    \begin{tabular}{ccccccc}
        \toprule
        && & \multicolumn{2}{c}{Relevance$\uparrow$} & Diversity$\downarrow$ \\
        \cmidrule(lr){4-5} \cmidrule(lr){6-6}
          \textbf{Model} & \#Param      & Method &  RL@10  &      Para@10  &  IntraRL \\\midrule
      \multirow{2}{*}{\includegraphics[width=18px]{img/llama.png}}&     \multirow{2}{*}{7B} & sampling &    0.183 & 0.487 &         0.308 \\
         &      &    batch &    0.230 & 0.469 &         0.314 \\
        \midrule %
         \multirow{4}{*}{\includegraphics[width=18px]{img/mistral.png}}&     \multirow{2}{*}{7B} & sampling &    0.251 & 0.523 &         0.322 \\
         &      &    batch &    0.229 & 0.449 &         0.346 \\
              \cmidrule(l){2-6}
         &    \multirow{2}{*}{8x7B} & sampling &    0.363 & 0.615 &         0.456 \\
         &     &    batch &    0.379 & 0.615 &         0.387 \\\midrule %
         \multirow{2}{*}{\includegraphics[width=18px]{img/gpt.png}} &     \multirow{2}{*}{n/a} & sampling &    0.544 & 0.719 &         0.545 \\
         &      &    batch &    0.422 & 0.629 &         0.433 \\
        \bottomrule
    \end{tabular}
    \caption{Performance of instruction suggestions. Relevance is measured by the highest ROUGE-L score (RL@10) and semantic similarity (Para@10) with the reference instruction, while diversity is measured by the Intra-RL score among the candidates.}
    \label{tab:suggest}
\end{table}

\subsection{Generation with Ambiguity Mitigation}
\label{sub:ex14}
To assess the practical effectiveness of our ambiguity mitigation framework, we conducted a final evaluation using LLM-generated additional instructions. 
Human annotators manually selected the most appropriate additional instruction $\hat{I}_{c_i}$ from $N$ options $\{\hat{I}_{c_i,j}\}_{j=1}^N$ generated in \S~\ref{sub:ex3}. 
The selected additional instruction was intended to facilitate the more accurate generation of the reference text $y_{\text{ref}}$. 
We then appended the best additional instructions across all categories to the initial instruction $I_{\text{init}}$, forming the refined instruction $\hat{I}_{\text{refined}}$ used for the downstream NLG task.

\paragraph{Settings}

We utilized additional instruction options generated by GPT-3.5 through sampling, as it demonstrated superior performance in \S~\ref{sub:ex3}. 
We randomly selected 100 test instances, resulting in a total of 2,140 additional instruction options. 
To evaluate the effectiveness of these refined instructions, we measured the similarity between the generated text $\hat{y}$ (produced using $\hat{I}_{\text{refined}}$) and the reference text $y_{\text{ref}}$, employing the ROUGE-L F1 score and BERTScore.

\paragraph{Results}

Incorporating LLM-generated additional instructions led to significant improvements: approximately 5.2-point increase in ROUGE-L (0.165 to \underline{0.217}) and a 4.6-point increase in BERTScore (0.273 to \underline{0.319}).\footnote{The underline denotes significant gains over baseline at $p < 0.05$.}
This demonstrates that LLM-generated instructions can significantly enhance the alignment of generated text with user expectations. 
Furthermore, we manually checked the outputs and found that in 94\% of cases where the quality of the output texts changed due to the additional instructions, the outputs more closely matched the reference texts.
These findings confirm that our framework for mitigating task ambiguity is effective in practical settings, highlighting its potential for real-world applications.

\section{Related Work}
\subsection{Ambiguity in NLP}
Ambiguity has long been a fundamental challenge in NLP~\cite{jurafsky1996probabilistic,carpuat-wu-2007-improving}, manifesting across a variety of tasks~\cite{min-etal-2020-ambigqa,pilault-etal-2023-interactive,bhaskar-etal-2023-benchmarking,liu-etal-2023-afraid}.
In this study, we specifically focused on \textit{task ambiguity}~\cite{finn2018probabilistic,NEURIPS2022_b43a0e8a,tamkin2023task} that arises when a model faces unclear and incomplete instructions or data.
Previous studies have addressed task ambiguities within the realm of natural language understanding (NLU)~\cite{finn2018probabilistic,NEURIPS2022_b43a0e8a,tamkin2023task}.
However, these approaches are insufficient for the complex and diverse context of NLG tasks, where mitigating ambiguity often requires more nuanced, instance-specific strategies.
 To address this gap, we tackle task ambiguity across a wide range of NLG tasks.

\subsection{Prompt Optimization}
Our study can also be positioned within the scope of prompt optimization, including techniques such as prompt paraphrasing~\cite{zhou2023large,pryzant-etal-2023-automatic,cho-etal-2023-discrete}
and detailed instruction integration~\cite{li2023guiding,bsharat2023principled,zhou2023instructionfollowing,wang2024promptagent}. 
We align with the latter approach by incorporating additional instructions to mitigate ambiguity in the initial prompts.
The primary distinction is that we uniquely focus on an \textit{instance-level} prompt optimization \textit{via a human-in-the-loop approach} for ambiguity mitigation, as opposed to the others' focus on optimizing a \textit{dataset-level} prompts or generating them \textit{automatically}.

\section{Conclusion}

We introduced AmbigNLG, a novel task designed to address the challenge of task ambiguity in instructions for NLG. 
We developed an ambiguity taxonomy that systematically categorizes types of ambiguities present in NLG instructions and proposed a method to refine initial instructions by providing clearer specifications.
We also constructed AmbigSNI$_\texttt{NLG}$ dataset, comprising 2,500 annotated instances, to facilitate the AmbigNLG task. 

Our comprehensive experiments with general LLMs demonstrated that our method significantly improves the alignment of generated text with user expectations. 
Furthermore, a case study involving real human interaction confirmed the practical utility of our approach.
These findings underscore the critical importance of addressing task ambiguity to fully harness the capabilities of LLMs in NLG tasks, paving the way for more precise and effective natural language interactions.

\section*{Acknowledgements}
We thank Yuki Arase from Tokyo Institute of Technology for her valuable feedback on this work. We are also thankful to Estevam Hruschka, Takuya Makino, and the other members of Megagon Labs for their insightful comments and suggestions.

\section*{Limitation}
While our proposed method effectively mitigates ambiguities based on the predefined taxonomy observed in the dataset, it currently does not address ambiguities that fall outside these categories. 
Extending our approach to encompass additional types of ambiguities would require systematizing other ambiguity categories and verifying their effectiveness.

In this study, we did not implement mechanisms to handle situations where the provided additional instructions might not fully meet user requirements. 
Recognizing this, incorporating mechanisms for iterative user interaction to refine instructions could further enhance the effectiveness of our approach. 

Moreover, when presenting multiple additional instructions to users, optimizing their selection through reranking could further enhance the effectiveness of the interaction. 
Developing methods to automatically select the more appropriate and promising additional instructions remains an open question.
Addressing this challenge could significantly improve user experience and the overall efficacy of ambiguity mitigation strategies.

\bibliography{custom}
\clearpage
\appendix

\section{Additional Details about Dataset Creation}
\subsection{Dataset Usage}\label{app:dataset_usage}

The AmbigSNI$_\texttt{NLG}$ dataset, with its ambiguity taxonomy and additional instructions, 
provides a foundation for research aimed at developing more reliable, efficient, and user-friendly NLG applications by mitigating the task ambiguity in NLG instructions.
Key uses of our dataset include:

\paragraph{Ambiguity Mitigation in NLG Tasks}

Indeed, by leveraging the taxonomy and additional instructions, developers and researchers can design systems that identify and mitigate ambiguities. 
This functionality is essential for generating more accurate and contextually relevant responses.

\paragraph{Instruction-Based NLG Model Training}

The dataset can be used to train models to interpret complex instructions that may contain ambiguities. 
This training helps models enhance their usability in real-world applications.

\paragraph{Request Clarification Model Development}

AmbigSNI$_\texttt{NLG}$ enables the development of models that can clarify users' requests when faced with ambiguous instructions. 
This functionality is vital for interactive systems that engage in dialogues with users to refine their requests, enhancing the overall effectiveness and user experience.

\paragraph{Benchmarking and Model Evaluation}

As a benchmark tool, the dataset enables an in-depth evaluation of how various NLG systems manage the task ambiguity in instructions.
Researchers can use the provided taxonomy and annotations to compare how different models address ambiguities, allowing for a detailed assessment of nuanced aspects of model performance.

\subsection{Preprocessing the SNI Benchmark \label{app:sampling_rule}}

The SNI benchmark comprises a wide variety of datasets, including both NLG and NLU datasets. 
For this study, we extracted only the NLG datasets from the SNI. We began by using the list of NLG datasets provided by~\citet{deb-etal-2022-boosting}.
We then refined this list by applying the following rules to clearly differentiate between NLG and NLU datasets.
A dataset qualifies as an NLG dataset only if it meets all the following criteria:
\begin{enumerate}
  \setlength{\parskip}{0cm} %
  \setlength{\itemsep}{0cm} %
    \item If the output text neither directly incorporates the input text nor the instruction. 
    \item If the output text consists of more than two words. 
    \item If the output is not composed solely of symbols or numbers.
\end{enumerate}

After completing this process, we renamed certain task names to more accurately reflect their content for our study, as detailed below:

\begin{itemize}[leftmargin=*]
\setlength{\parskip}{0cm} %
  \setlength{\itemsep}{0cm} %
    \item Question answering $\rightarrow$ Long-form question answering (QA)
    \item Information extraction $\rightarrow$ Attribute Generation
    \item Named Entity Recognition $\rightarrow$ Generation-based Named Entity Recognition (NER)
    \item Keyword Tagging $\rightarrow$ Keyword Generation
    \item Overlap Extraction $\rightarrow$ Generation-based Overlap Extraction (OE)
\end{itemize}

\subsection{Annotation Step in Curation\label{app:curation}}
In the curation process in \S\ref{sub:annotation}, we ensured the quality of the additional instructions through a three-step process:
\begin{enumerate}[noitemsep, nolistsep]
    \item An author crafted additional instructions for the sampled instances, following the same guidelines used to fine-tune GPT-3.5, as outlined in Table~\ref{tab:category_prompt}.
    \item The same author then carefully refined these instructions, ensuring that:
    \begin{itemize}[noitemsep, nolistsep]
        \item The content remained consistently relevant
        \item No explicit answers were included within the additional instructions
        \item There was no content overlap with additional instructions for other ambiguity categories
        \item There was no content overlap between the additional instructions and the initial instructions or input text
    \end{itemize}
    \item Other authors reviewed and revised the additional instructions as necessary.
\end{enumerate}

\subsection{Rule-based Annotation\label{app:rule_based_annotation}}
Additional instructions for the \textsc{Keyword} and \textsc{Length} categories can be derived solely from the output text based on predefined rules, without an LLM.
The annotation process for each is as follows:

\paragraph{\textsc{Keyword}}
 We utilize the lightweight unsupervised keyword extraction method Yake~\cite{CAMPOS2020257} to extract the Top-$n$ most significant keywords or key phrases from the output text.
These extracted keywords or key phrases are then used to fill the template `Include \_\_\_ in your response.'
However, selecting an excessively high value of $n$ can result in an impractical setup. 
Therefore, we define $n$ based on the output length, ensuring that only a reasonable number of keywords or key phrases are provided.
\[
n = \max \left\{ m \,|\, m \leq 4, \sum_{i=1}^{m} w_i \leq 0.4 \cdot W \right\}
\]
where $W$ is the total word count in the output text and $w_i$ is the word count in the $i$-th key phrase.

\paragraph{\textsc{Length}}
Using NLTK~\cite{bird-2006-nltk}, we extract the word count $n$ from the output text and fill in the template `Answer with \_\_\_ words' accordingly.
However, configuring an LLM to generate exactly $n$ words is impractical.
Instead of specifying an exact count, we define a range using the phrase `$a$ to $b$ words.'

\[
(a, b) = \left(\left\lfloor \frac{n}{10} \right\rfloor \times 10, \left(\left\lfloor \frac{n}{10} \right\rfloor + 1\right) \times 10\right)
\]
In situations where $n$ is 10 or less, we modify the template to use the phrase `less than $b$ words.'

\subsection{Examples from AmbigSNI$_\texttt{NLG}$ dataset\label{app:examples}}

Table~\ref{tab:example1} and \ref{tab:example2} present the examples from the AmbigSNI$_\texttt{NLG}$ dataset, illustrating the instruction, input text, reference text, assigned ambiguity category, and the corresponding additional instruction for the category.

\begin{table*}[]
    \centering
    \begin{tabular}{cp{11cm}}\toprule
    Instruction & In this task, we ask you convert a data table of restaurant descriptions into fluent natural-sounding English sentences. The input is a string of key-value pairs; the output should be a natural and grammatical English sentence containing all the information from the input.\\
    Input     & name[The Golden Palace], \textcolor{forestgreen}{eatType[coffee shop]}, \textcolor{forestgreen}{food[Indian]}, \textcolor{softred}{priceRange[moderate]}, \textcolor{softred}{customer rating[3 out of 5]}, \textcolor{forestgreen}{area[city centre]} \\
    Reference     &\textcolor{forestgreen}{The Golden Palace is a coffee shop in city centre that serves Indian food.} \textcolor{softred}{It has a customer rating of 3 out of 5 and moderately priced.} \\
    Ambiguity categories & \textsc{Planning} \\
    Additional instruction &Please generate the output based on the following outline: 1. \textcolor{forestgreen}{Description and location of The Golden Palace} 2. \textcolor{softred}{Customer rating and pricing of The Golden Palace}\\\bottomrule
    \end{tabular}
    \caption{Example 1 (id: \texttt{task957-75dd6eba92a649ba81524c3a0594d57c}) from AmbigSNI$_\texttt{NLG}$ dataset. The input table contains multiple contents, making it ambiguous in the initial instructions how each content should be represented in the output. Therefore, an additional instruction regarding  \textsc{Planning} was assigned to specify that the  \textcolor{softred}{customer ratings and pricing} should be explained after \textcolor{forestgreen}{describing the restaurant's information}.}
    \label{tab:example1}
\end{table*}

\begin{table*}[]
    \centering
    \begin{tabular}{cp{11cm}}\toprule
    Instruction & In this task, you are given an article. Your task is to summarize the article in a sentence.\\
    Input     &  \textcolor{forestgreen}{Aslef and RMT members are due to walk out for six days from 9 January.} \textcolor{softred}{The Confederation of Passenger Transport (CPT) said bus operators from Cornwall to Northumberland were ready to send vehicles to the South East. Southern said it was still deciding what services might be offered. A CPT spokeswoman said: ``We have had a very good response from quite a few members.'' It has sent Southern's parent company, the Go-Ahead Group, a list of operators including family-run firms which are ready to provide buses. Southern said it planned to announce on Wednesday what rail replacement services might be offered ``to some commuters'' but warned there would be no trains at all during the strike.} \textcolor{oceanblue}{Three weeks ago the government said officials were liaising with CPT ``to determine how bus and coach operators can best assist with providing alternative transport''. BBC South East understands the Army was asked before Christmas to prepare contingency plans for soldiers to drive buses.} ... \\
    Reference     & \textcolor{softred}{Dozens of bus and coach companies across England have offered vehicles for rail replacement services during the next Southern train drivers' strike.}\\
    Ambiguity categories &\textsc{Theme} \\
    Additional instruction &Primarily discuss the following theme: \textcolor{softred}{Provision of alternative transport during a train drivers' strike.}\\\bottomrule
    \end{tabular}
    \caption{Example 2 (id: \texttt{task1290-643d125a902345fca21b2c8a83ff4006}) from AmbigSNI$_\texttt{NLG}$ dataset. The input article includes multiple sub-themes, such as \textcolor{forestgreen}{strike schedules},  \textcolor{softred}{alternative transportation}, and \textcolor{oceanblue}{government collaboration}, making it ambiguous which theme should be focused on in the summary. Therefore, an additional instruction regarding the \textsc{Theme} was assigned to specify focusing on \textcolor{softred}{alternative transportation}. }
    \label{tab:example2}
\end{table*}

\subsection{Further Statistics of Additional Instruction \label{app:cost}}
\paragraph{Sequence Length}

\begin{figure}[t]
    \centering
    \includegraphics[width=\linewidth]{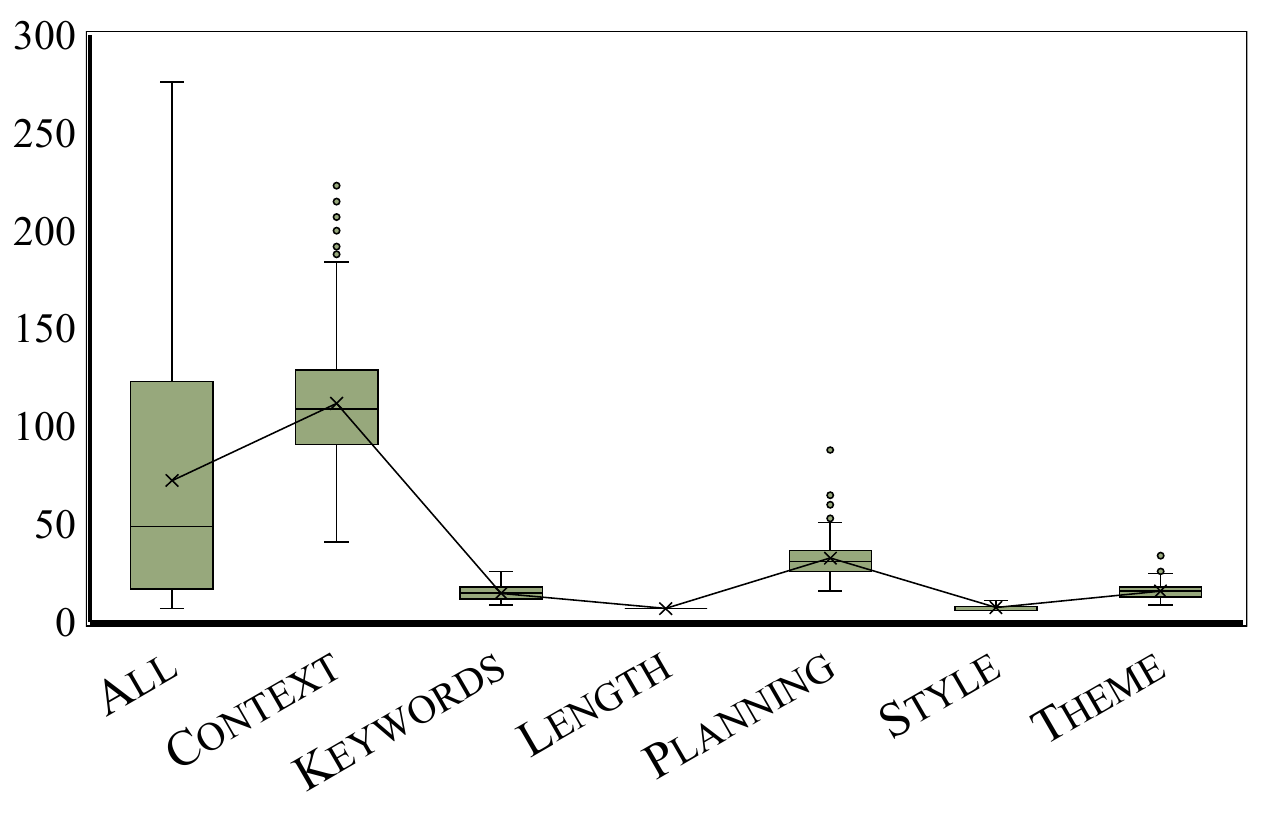}
    \caption{Length distribution of the additional instruction.}
    \label{fig:length_dist}
\end{figure}

We display the length distribution of additional instruction for each ambiguity category in Figure~\ref{fig:length_dist}.
The sequence length of the concatenated additional instructions (All), which encompass all assigned ambiguity categories, averages 49 words, with a maximum of 276 words.
The length varies significantly depending on the assigned ambiguity category, tending to be longer when \textsc{Context} is included, as this category typically results in the longest sequence length.

\paragraph{Minimized Information Leakage}

We confirmed that information leakage of the reference text is minimized by enforcing a constraint on the prompt (in Table~\ref{tab:category_prompt}) to ensure that the answer itself is not included in the additional instruction $\hat{I}_{c}$.
To validate this, we assessed the overlap between $\hat{I}{c}$ and $y_{\text{ref}}$ using the ROUGE score, which resulted in a score of 0.177. 
This is notably lower than the ROUGE score of 0.229 between input text $x$ and reference text $y_{\text{ref}}$, indicating the effectiveness of the constraint.

\section{Further Experimental Details\label{app:further_analysis}}
\subsection{Computational Details}
We performed all experiments to run on eight 80GB A100 GPUs.
For the open-sourced LLMs, we used vLLM~\cite{Kkwon2023vllm}, which implements a variety of efficiency tricks for the transformer model to make the LLMs' inference faster~\cite{pope2023efficiently,dao2022flashattention}.
For the proprietary LLMs, we used the official OpenAI library to call the API.

\subsection{Results about Ambiguity Mitigation}

\paragraph{Additional Cost by the Concatenation}
Our mitigation method involves augmenting the initial instruction with the additional instruction, which increases the sequence length.
To quantify the cost, we use the OpenAI API as an example, which represents the highest-cost option in our experiments.
 Using the \texttt{gpt-3.5-turbo} model at \$0.0005 per 1,000 tokens, the average additional cost per instance is \$0.0000245, with a maximum of \$0.000138.
For the more expensive \texttt{gpt-4-32k} model, priced at \$0.06 per 1,000 tokens, the average additional cost per instance rises to \$0.00294 and a maximum of \$0.01656.
These results indicate that the proposed framework enhances performance while incurring only minimal additional costs.

\paragraph{Results about instruction following}
To determine whether additional instructions make instruction too complex for LLMs to follow, we evaluated the Instruction Following (IF) score for models both without mitigation (using the initial instructions $I_{\text{init}}$) and with mitigation (using the refined instructions $I_{\text{refined}}$). 
Similar to \citep{liu-etal-2023-g}, we employed GPT-4 as the evaluator, utilizing a five-point scale. 
We randomly selected 100 instances for this analysis. 
The results, shown in Table~\ref{tab:if_score}, indicate that the IF scores for $I_{\text{refined}}$ consistently exceeded those for $I_{\text{init}}$. 
This suggests that our additional instructions do not overcomplicate the refined instruction. 
We hypothesize that the higher IF scores with the refined instructions are due to the clearer and more specific criteria they provide, which enhance the models' ability to follow instructions accurately.

\begin{table}[]
    \centering
    \resizebox{\linewidth}{!}{
    \small
    \begin{tabular}{crrrr}\toprule
   Instruction& Llama-2&	Mistral	&Mixtral&	GPT-3.5\\\midrule
    $I_{\text{init}}$   & 3.87&	4.41	&4.46	&4.12 \\
     $I_{\text{refined}}$  &\textbf{4.15}&	\textbf{4.59}	&\textbf{4.70}&	\textbf{4.43}\\\bottomrule
    \end{tabular}
    }
    \caption{Instruction following (IF) score. }
    \label{tab:if_score}
\end{table}

\subsection{Results about Ambiguous Category Identification}
\paragraph{Overall Results}
We display the results for each taxonomy in \S\ref{sub:ex3} in Table~\ref{tab:all_ex2}.
\begin{table*}[t]
    \centering
    \small
    \setlength\tabcolsep{3pt}
    \begin{tabular}{ccccccccccccccc}
    \toprule
    & &        & \multicolumn{4}{c}{\textsc{Context}} & \multicolumn{4}{c}{\textsc{Keywords}} & \multicolumn{4}{c}{\textsc{Length}} \\
    \cmidrule(lr){4-7} \cmidrule(lr){8-11} \cmidrule(lr){12-15}
    \textbf{Model} & \#Param  &      ICL  &     TPR &   TNR & B-Acc &   Acc &      TPR &   TNR & B-Acc &   Acc &    TPR &   TNR & B-Acc &   Acc \\
    \midrule
    \multirow{2}{*}{\includegraphics[width=15px]{img/llama.png}}& 7B & \xmark &   98.12 &  2.45 & 50.29 & 35.60 &    97.15 &  1.87 & 49.51 & 38.70 &  98.65 &  1.84 & 50.24 & 19.75 \\
    & 7B & \cmark &   12.70 & 88.75 & 50.73 & 62.40 &    10.09 & 85.25 & 47.67 & 56.20 &  12.97 & 83.87 & 48.42 & 70.75   \\\midrule
    \multirow{4}{*}{\includegraphics[width=15px]{img/mistral.png}}& 7B & \xmark &   99.86 &  0.23 & 50.04 & 34.75 &   100.00 &  0.08 & 50.04 & 38.70 &  99.73 &  0.12 & 49.93 & 18.55\\
    & 7B & \cmark &   55.41 & 52.95 & 54.18 & 53.80 &    57.44 & 51.75 & 54.60 & 53.95 &  55.68 & 49.69 & 52.68 & 50.80 \\
    \cmidrule(l){2-15}
    & 8x7B & \xmark &   90.04 & 11.17 & 50.61 & 38.50 &    98.97 &  0.90 & 49.93 & 38.80 &  98.11 &  4.05 & 51.08 & 21.45 \\
    & 8x7B & \cmark &   19.19 & 84.62 & 51.91 & 61.95 &    25.36 & 79.22 & 52.29 & 58.40 &  17.30 & 85.09 & 51.19 & 72.55  \\\midrule
    \multirow{2}{*}{\includegraphics[width=15px]{img/gpt.png}} & n/a & \xmark &   68.83 & 35.58 & 52.20 & 47.10 &    80.47 & 23.96 & 52.21 & 45.80 &  84.32 & 21.10 & 52.71 & 32.80 \\
    & n/a & \cmark &   84.70 & 17.83 & 51.27 & 41.00 &    89.39 &  8.96 & 49.18 & 40.05 &  90.54 &  7.24 & 48.89 & 22.65\\\toprule
        & &        & \multicolumn{4}{c}{\textsc{Planning}} & \multicolumn{4}{c}{\textsc{Style}} & \multicolumn{4}{c}{\textsc{Theme}} \\
    \cmidrule(lr){4-7} \cmidrule(lr){8-11} \cmidrule(lr){12-15}
    \textbf{Model} & \#Param  &      ICL  &     TPR &   TNR & B-Acc &   Acc &      TPR &   TNR & B-Acc &   Acc &    TPR &TNR & B-Acc &   Acc\\
    \midrule
    \multirow{2}{*}{\includegraphics[width=15px]{img/llama.png}}& 7B & \xmark &    99.15 &  1.49 & 50.32 &  7.25 & 100.00 &  1.63 & 50.81 &  3.15 &  96.75 &  1.66 & 49.21 & 28.00 \\
    & 7B & \cmark &      21.19 & 88.10 & 54.64 & 84.15 &  19.35 & 86.24 & 52.80 & 85.20 &   7.76 & 88.66&48.21& 66.25\\\midrule
    \multirow{4}{*}{\includegraphics[width=15px]{img/mistral.png}}& 7B & \xmark &     100.00 &  0.32 & 50.16 &  6.20 & 100.00 &  0.10 & 50.05 &  1.65 & 100.00 &  0.07 & 50.03 & 27.75 \\
    & 7B & \cmark &       48.31 & 47.02 & 47.66 & 47.10 &  67.74 & 50.89 & 59.32 & 51.15 &  45.49 & 43.57 & 44.53 & 44.10 \\
    \cmidrule(l){2-15}
    & 8x7B & \xmark &      94.07 &  4.14 & 49.11 &  9.45 & 100.00 &  3.30 & 51.65 &  4.80 &  98.38 &  2.70 & 50.54 & 29.20 \\
    & 8x7B & \cmark &   11.86 & 84.96 & 48.41 & 80.65 &  25.81 & 85.68 & 55.74 & 84.75 &  21.30 & 77.52 & 49.41 & 61.95 \\\midrule
    \multirow{2}{*}{\includegraphics[width=15px]{img/gpt.png}} & - & \xmark &    76.27 & 20.56 & 48.42 & 23.85 &  74.19 & 17.17 & 45.68 & 18.05 &  93.86 & 19.64 & 56.75 & 40.20 \\
    & - & \cmark &     85.59 &  9.78 & 47.69 & 14.25 &  87.10 & 10.87 & 48.98 & 12.05 &  87.55 & 10.10 & 48.82 & 31.55\\
    \bottomrule
    \end{tabular}
    \caption{Overall results of category identification in \S\ref{sub:ex2}.}\label{tab:all_ex2}
    \end{table*}

\subsection{Results about the Instruction Suggestion}
\paragraph{Further Results}
In Section~\ref{sub:ex3}, we employed an approach that fills in templates when suggesting additional instructions. 
Here, for comparison, we examine the results of an open-ended approach where additional instructions are generated without using templates.
Table~\ref{tab:suggest_open_end} showcases that open-ended generation is more diverse because it doesn't follow a single template to generate suggestions, but generally less relevant than the fill-in-the-blank approach (Table~\ref{tab:suggest}).
Therefore, we adopted the template-based approach in the main experiment.

\begin{table}[t]
    \centering
    \small
    \setlength{\tabcolsep}{3pt}
    \begin{tabular}{ccccccc}
        \toprule
         & & & \multicolumn{2}{c}{Relevance$\uparrow$} & Diversity$\downarrow$ \\
        \cmidrule(lr){4-5} \cmidrule(lr){6-6}
         \textbf{Model} & \#Param &   Method &  RL@10  &      Para@10  &  IntraRL \\\midrule
         \multirow{2}{*}{\includegraphics[width=18px]{img/llama.png}} &     \multirow{2}{*}{7B} & sampling &         0.128 & 0.454 &         0.313 \\
         &      &    batch &         0.165 & 0.440 &         0.306 \\
        \midrule %
         \multirow{4}{*}{\includegraphics[width=18px]{img/mistral.png}} &     \multirow{2}{*}{7B} & sampling &         0.152 & 0.455 &         0.313 \\
         &     &    batch &         0.171 & 0.400 &         0.345 \\
              \cmidrule(l){2-6}
         &    \multirow{2}{*}{8x7B} & sampling &         0.183 & 0.516 &         0.370 \\
         &     &    batch &         0.215 & 0.517 &         0.284 \\\midrule %
         \multirow{2}{*}{\includegraphics[width=18px]{img/gpt.png}}  &     \multirow{2}{*}{n/a} & sampling &         0.216 & 0.544 &         0.384 \\
         &      &    batch &         0.201 & 0.508 &         0.286 \\
        \bottomrule
    \end{tabular}
    \caption{Instruction suggestions performance generated in a open-ended manner.}
    \label{tab:suggest_open_end}
\end{table}

\begin{table*}[]
    \centering
    \small
    \begin{tabular}{lrrrrrr}
    \toprule
  \textbf{Tasks} &  \textsc{Context}	& \textsc{Keywords}	 &\textsc{Length}	& \textsc{Planning}&	 \textsc{Style}	 &\textsc{Theme}\\\toprule
       Question Generation&	0.272&	0.356	&0.154	&0.068&	0.002	&0.480\\
Long-form QA	&0.443	&0.260	&0.347&	0.027&	0.015	&0.122\\
Sentence Composition&	0.347	&0.427	&0.147	&-&	0.027	&0.196\\
Title Generation	&0.388	&0.454&	0.098	&0.011	&0.049	&0.585\\
Attribute Generation&	0.424	&0.223&	0.216	&-&	-&	0.108\\
Text Completion&	0.373	&0.814&	0.280&	-&	0.017&	0.195\\
Data to Text	&0.255	&0.518&	0.318&	0.145	&0.064&	0.273\\
Question Rewriting&	0.391&	0.245	&0.082&	0.055&	-	&0.200\\
Wrong Candidate Generation&	0.227&	0.173&	0.145&	-	&0.009&	0.073\\
Story Composition&	0.356&	0.713&	0.257	&0.139	&0.040&	0.317\\
Summarization&	0.343	&0.505	&0.263&	0.222	&0.010&	0.354\\
Code to Text	&0.224&	0.079&	0.105&	0.197&	-&	-\\
Dialogue Generation&	0.382&	0.605&	0.263&	0.276	&0.026	&0.329\\
Generation-based NER	&0.358&	0.113&	0.057&	-	&-	&0.057\\
Paraphrasing	&0.625&	0.200&	0.050	&-&	0.025&	0.075\\
Sentence Perturbation&	0.545&	0.636	&-	&0.030	&-	&0.121\\
Explanation&	0.500	&0.591	&0.455	&0.136	&-	&0.091\\
Fill in The Blank&	0.895&	0.368&	0.105&	0.053	&0.053	&0.211\\
Question Decomposition	&0.333&	0.600&	0.133	&-&	-&	0.067\\
Grammar Error Correction	&0.154&	0.154	&-	&-	&-	&-\\
Text Simplification&	0.308&	0.231	&-	&0.077&	0.077&	0.308\\
Sentence Compression	&0.100	&0.500	&0.200	&0.100	&0.100&	0.100\\
Keyword Generation&	0.333	&0.111&	0.222	&-&	-&	0.222\\
Poem Generation&	0.444&	0.889	&0.222&	-	&-&	0.222\\
Sentence Expansion&	0.286	&1.000	&0.143&	-	&-	&-\\
Number Conversion	&-	&-&	-	&-	&-&	-\\
Entity Generation&	1.000	&-&	-	&-&	-&	0.333\\
Style Transfer&	0.500	&0.500&	-	&-	&-	&-\\
Generation-based OE &	1.000&	1.000	&-&	-&	-	&-\\\bottomrule
    \end{tabular}
    \caption{Overall results of ambiguity mitigation across all tasks in \S\ref{sub:ex1}. `-' indicates that the category is not assigned to the instances in the task.}
    \label{tab:all_tasks}
\end{table*}

\section{List of prompts\label{sec:list_of_prompts}}

\begin{table*}[th]
    \centering
    \begin{tabular}{cp{13cm}}\toprule
    \multicolumn{2}{c}{\textbf{System Message}}\\\toprule
    &You are an AI assistant addressing various ambiguities in NLP task instructions. Your role involves complementing incomplete information by filling in the blanks within the provided template. The template you have filled in is used as the additional instruction.\\\toprule
    \multicolumn{2}{c}{\textbf{User Message}} \\\toprule
    \textbf{Taxonomy} & Prompt\\\toprule
    \textsc{Context} &   Please identify what additional context, such as background or external knowledge, will encourage the accurate generation from input to output text. Subsequently, write a concise paragraph containing the required context and other related content. Fill in the blank of the following template: 'Additional context: {paragraph}'. Ensure that it's not clear which part of the paragraph corresponds to the output text. Please answer with the additional context needed to solve the task, not the solution to the task itself.\\\hdashline
        \textsc{Keywords}&	-\\\hdashline
        \textsc{Length}&	-\\\hdashline
    \textsc{Planning}&	Please describe the output text structure by listing a concise topic for each sentence. Fill in the blank of the following template: 'Please generate the output based on the following outline: 1. {topic1} 2. {topic2} ...'. Ensure that the number of items in the list matches the number of sentences in the output text. Make sure the response is brief and generalized, not detailed.\\\hdashline
    \textsc{Style}&Please select the writing style of the output text from the following options: descriptive, expository, narrative, persuasive, directive, conversational, technical, journalistic, review, poetic, formal, informal, optimistic, assertive, dramatic, humorous, sad, passive-aggressive, worried, friendly, curious, encouraging, surprised, cooperative. Fill in the blank of the following template: 'Write in a {style} style.'. You are allowed to select multiple styles if necessary. If none of the styles align with the text, please respond with 'neutral'\\\hdashline
    \textsc{Theme}&Please identify the single, most dominant content of the output text and provide a clear and succinct description of it. Fill in the blank of the following template: 'Primarily discuss the following theme: {theme}'. Make sure the response is brief and generalized, not detailed. Concentrate on the theme of the output text, rather than on the input text, instruction, or the overall task. The reply may contain hints of the output text, but should refrain from encapsulating its full content.\\
    \bottomrule
    \end{tabular}
    \caption{Category prompts for fill-in-the-blank in dataset creation. (For \textsc{Keywords} and \textsc{Length}, we adopted the rule-based annotation as described in \S\ref{app:rule_based_annotation}.)\label{tab:category_prompt}}
\end{table*}

\begin{tcolorbox}[fontupper=\ttfamily, title={\small Prompts for annotation used in \S\ref{sub:annotation}. For \textsc{Context} }, fonttitle=\hypersetup{linkcolor=white,urlcolor=white}]
\scriptsize
    \{Category prompt in Table~\ref{tab:category_prompt}\}\\
    \\
    \# Instruction\\
    \{instruction\}\\
    \\
    \# Input text:\\
    \{input\}\\
    \\
    \# Output text:\\
    \{output\}\\
    \\
    \# Template:\\
\end{tcolorbox}

\begin{tcolorbox}[fontupper=\ttfamily, title={\small Prompts for annotation used in \S\ref{sub:annotation}. For \textsc{Theme}, \textsc{Planning}, \textsc{Style} }, fonttitle=\hypersetup{linkcolor=white,urlcolor=white}]
\scriptsize
    \{Category prompt in Table~\ref{tab:category_prompt}\}\\
    \\
    \# Task Category\\
    \{task category\}\\
    \\
    \# Input text:\\
    \{input\}\\
    \\
    \# Output text:\\
    \{output\}\\
    \\
    \# Template:\\
\end{tcolorbox}

   \begin{tcolorbox}[fontupper=\ttfamily, title={\small Prompt for validation (\underline{\textbf{Clarity}}) used in \S\ref{validation}. }, fonttitle=\hypersetup{linkcolor=white,urlcolor=white}]
\scriptsize
    \# Instruction\\
\{instruction\}\\
\\
For the instruction above, please assess that combining the additional instruction below with the instruction either increases, decreases, or maintains the ambiguity level in the instruction to lead the precise generation of output text from the input text.\\
More specifically, focus on the aspect of ‘\{ambiguity category\}’ (\{description\}).\\
Answer with ‘More ambiguous’, ‘Less ambiguous’, or ‘Unchanged’.\\
\\
\# Input text:\\
\{input\}\\
\\
\# Output text:\\
\{output\}\\
\\
\# additional instruction:\\
\{additional instruction\}\\
\\
\# Answer:\\
\end{tcolorbox}

   \begin{tcolorbox}[fontupper=\ttfamily, title={\small Prompt for  validation (\underline{\textbf{Utility}}) used in \S\ref{validation}.}, fonttitle=\hypersetup{linkcolor=white,urlcolor=white}]
\scriptsize
       Below is an input text that provides further context, paired with an instruction that describes a task.\\
    Write a response that appropriately completes the request.\\
    \\
    \# Input text:\\
    \{input\}\\
    \\
    \# Instruction:\\
    \{instruction\}\\
    \\
    \# Response:\\
    \end{tcolorbox}

\begin{table*}[t]
    \centering
    \begin{tabular}{cp{13cm}}\toprule
    \textbf{Taxonomy}&Task Definition\\\toprule
    \textsc{Length} & Length: Opt for this category if the instruction does not provide specifics about the desired length of the output, whether in terms of words or sentences. Clearing this ambiguity will lead to a more precise length output.\\
    \textsc{Keywords} & Keyword: Select this category if the instruction does not mention specific keywords to be used in the output text. Resolving this ambiguity will ensure that the necessary keywords are incorporated in the output.\\
    \textsc{Context} & Context: Choose this category if the instruction lacks the required context information, such as background or external knowledge crucial for task completion. Resolving this ambiguity will provide the crucial context for the task.\\
    \textsc{Theme} & Theme: Choose this category if the instruction does not clearly define the specific theme to be discussed in the output text. Clearing this ambiguity will provide a clear direction for the output.\\
    \textsc{Planning} & Plan: Select this category if the instructions doesn't provide guidance on content planning for the output document. Resolving this ambiguity will result in the desired structured output.\\
    \textsc{Style} & Style: Choose this category if the instruction does not specify the style of the output text. Clearing this ambiguity will ensure that the output aligns with the desired style.\\
    \textsc{None} & None: Choose this category if the instructions are clear, define all aspects of the task well, and lead to a nearly single output.
 \\\bottomrule
        \end{tabular}
    \caption{Prompt for ambiguity identification used in \S\ref{sub:ex2}.}
    \label{tab:ex2_task_definition}
\end{table*}

\begin{tcolorbox}[fontupper=\ttfamily, title={\small Prompt for executing downstream tasks used in \S\ref{sub:ex1} and \S\ref{sub:ex14}}, fonttitle=\hypersetup{linkcolor=white,urlcolor=white}]
\scriptsize
       Below is an input text that provides further context, paired with an instruction that describes a task.\\
Provide a direct response that appropriately completes the request without additional explanations or details.\\
    \\
    \# Input text:\\
    \{input\}\\
    \\
    \# Instruction:\\
    \{instruction\}\\
    \\
    \# Response:\\
\end{tcolorbox}

\begin{tcolorbox}[fontupper=\ttfamily, title={\small Prompt for ambiguity identification used in \S\ref{sub:ex2}}, fonttitle=\hypersetup{linkcolor=white,urlcolor=white}]
\scriptsize
   Your task involves identifying the category of ambiguity in the given instruction to generate output text from the given input text. \\
    Ambiguity in instruction means that there are several possible output texts from the single input text.\\
    On the other hand, when the ambiguity is clarified, the task becomes straightforward, leading to a nearly single output.\\
    Here are the available categories: \{category list\}.\\

     \{task\_definition in Table~\ref{tab:ex2_task_definition}\}\\

If there are multiple ambiguities, please provide your answer as a comma-separated list.\\

\# Instruction:\\
\{instruction\}\\

\# Input text:\\
\{input\}\\

\# Response:\\
\end{tcolorbox}

\begin{tcolorbox}[fontupper=\ttfamily, title={\small Prompt for instruction suggestion used in \S\ref{sub:ex3}}, fonttitle=\hypersetup{linkcolor=white,urlcolor=white}]
\scriptsize
 To resolve the specified ambiguity in the instruction, provide an additional specific instruction by infilling the provided template. Ensure this added information aligns with the primary objective of the task, supports understanding of complex concepts, or aids in narrowing down the scope to generate more precise responses.\\

\# Input Text:\\
\{input\_text\}\\

\# Instruction:\\
\{instruction\}\\

\# Ambiguity to Resolve:\\
\{ambiguity\_category\}: \{ambiguity\_definition\}\\

\# Template to Infill:\\
\{template\}\\

\# Additional Instruction:
\end{tcolorbox}

\begin{tcolorbox}[fontupper=\ttfamily, title={\small G-Eval Prompt for instruction following evaluation used in \S\ref{sub:ex1}}, fonttitle=\hypersetup{linkcolor=white,urlcolor=white}]
\scriptsize
      Below is an instruction for evaluating the instruction-following ability of a language model in the context of generating text based on specific instructions. The evaluation ranges from 1 to 5, with 1 being the lowest and 5 the highest in terms of accuracy and adherence to the given instruction. If there are parts of the task instructions enclosed in asterisks (*), please focus your evaluation particularly on whether it adheres to those highlighted sections. \\
      \\
      \# Evaluation Criteria: \\
      1. The output is unrelated to the given instruction.  \\
      2. The output vaguely relates to the instruction but misses key elements.  \\
      3. The output is somewhat accurate but lacks detail or has minor inaccuracies.  \\
      4. The output is accurate and detailed, with only negligible issues.  \\
      5. The output perfectly matches the instruction with high accuracy and detail.  \\
       \\
       \# Instruction:  \\
       \{instruction\}  \\
      \*\{additional instruction\}\*  \\
        \\
        \# Input Text:  \\
        \{input text\}  \\
         \\
         \# Output Text:  \\
         \{output text\}  \\
          \\
          \# Evaluation Form (scores ONLY):\\
\end{tcolorbox}

\clearpage

\end{document}